\documentclass{article} % For LaTeX2e
\usepackage{iclr2025_delta,times}

\usepackage{amsmath,amsfonts,bm}

\def\eqref#1{equation~\ref{#1}}
\def\1{\bm{1}}

\def\rc{{\textnormal{c}}}

\def\rx{{\textnormal{x}}}
\def\ry{{\textnormal{y}}}
\def\rz{{\textnormal{z}}}

\def\rvy{{\mathbf{y}}}

\DeclareMathAlphabet{\mathsfit}{\encodingdefault}{\sfdefault}{m}{sl}
\SetMathAlphabet{\mathsfit}{bold}{\encodingdefault}{\sfdefault}{bx}{n}

\usepackage{hyperref}
\usepackage{graphicx}
\usepackage{float} 
\usepackage{url}

\usepackage{tikz}
\usepackage{algorithm}
\usepackage{algpseudocode}
\usetikzlibrary{shapes.geometric, arrows}
\usepackage{chngcntr}
 
\tikzstyle{block} = [rectangle, draw, , text centered]
\tikzstyle{arrow} = [thick,->,>=stealth]
\usepackage[inkscapelatex=false]{svg}

\title{Measuring Semantic Information Production in Generative Diffusion Models}

\author{Florian Handke\\
Department of Artificial Intelligence\\
Radboud University\\
\texttt{florian.handke@ru.nl} \\
\And
Félix Koulischer  \\
Department of Information Technology \\
Ghent University \\
\texttt{felix.koulischer@ugent.be} \\
\And
Gabriel Raya\\
Jheronimus Academy of Data Science\\
Tilburg University \\
\texttt{g.raya@jads.nl}\\
\And
Luca Ambrogioni \\
Donders Institute for Brain, Cognition and Behaviour \\
Radboud University \\
\texttt{l.ambrogioni@donders.ru.nl}
}

\iclrfinalcopy % Uncomment for camera-ready version, but NOT for submission.
\begin{document}
\maketitle
\begin{abstract}
It is well known that semantic and structural features of the generated images emerge at different times during the reverse dynamics of diffusion, a phenomenon that has been connected to physical phase transitions in magnets and other materials. In this paper, we introduce a general information-theoretic approach to measure when these class-semantic "decisions" are made during the generative process. By using an online formula for the optimal Bayesian classifier, we estimate the conditional entropy of the class label given the noisy state. We then determine the time intervals corresponding to the highest information transfer between noisy states and class labels using the time derivative of the conditional entropy. We demonstrate our method on one-dimensional Gaussian mixture models and on DDPM models trained on the CIFAR10 dataset. As expected, we find that the semantic information transfer is highest in the intermediate stages of diffusion while vanishing during the final stages. However, we found sizable differences between the entropy rate profiles of different classes, suggesting that different "semantic decisions" are located at different intermediate times. 
\end{abstract}
\section{Introduction} % 1.25 pages (figure of symmetry breaking in 1D 2 delta model and conditional entropy derivative)
Generative diffusion models \citep{sohldickstein2015deepunsupervisedlearningusing} operate through two complementary processes: a predefined forward stochastic process, which gradually adds noise to data, and its correspondingly learned reverse process, which can be interpreted as a form of dynamic denoising. In the continuous limit, both processes can be described in terms of differential equations \citep{song2021scorebasedgenerativemodelingstochastic}, i.e., diffusion processes. The reverse dynamics of this generative process has been studied extensively in recent years, revealing profound theoretical connections to key concepts in the statistical physics of phase transitions. Specifically, \citet{raya2023spontaneoussymmetrybreakinggenerative} and \citet{ambrogioni2024statisticalthermodynamicsgenerativediffusion} have provided theoretical and empirical evidence for the existence of spontaneous symmetry-breaking phase transitions during generation, which correspond to temporally localized "decision windows" that align with bifurcations in the generative process. In line with their findings, \citet{Biroli_2024} have characterized a \textit{speciation} phase
during which a sample's general features are determined, which is consistent with the observation that class-specific characteristics emerge at different times during generation \citep{sclocchi2024phasetransitiondiffusionmodels}. This phenomenon was further studied mathematically by \citet{li2024criticalwindowsnonasymptotictheory}, who have provided a theoretical framework by which the bounds of this \textit{critical window} can be predicted, yet for a restricted set of initial distributions. From a more applied side, \citet{kynkäänniemi2024applyingguidancelimitedinterval} have demonstrated that restricting classifier-free guidance to an intermediate range of noise levels is essential in preserving both sample diversity and quality, emphasizing the practical importance of a critical interval in conditional generation. 

In spite of this growing body of theoretical literature, a practical and scalable method to localize these decision windows in models trained on complex datasets with unknown distributions is still missing. In this paper, using ideas from information theory, we introduce a distribution-agnostic method to localize the temporal windows corresponding to class-semantic structure generation in trained diffusion models. 

% figure 1
\includesvg[width=1\linewidth]{figures/fig_1.svg}

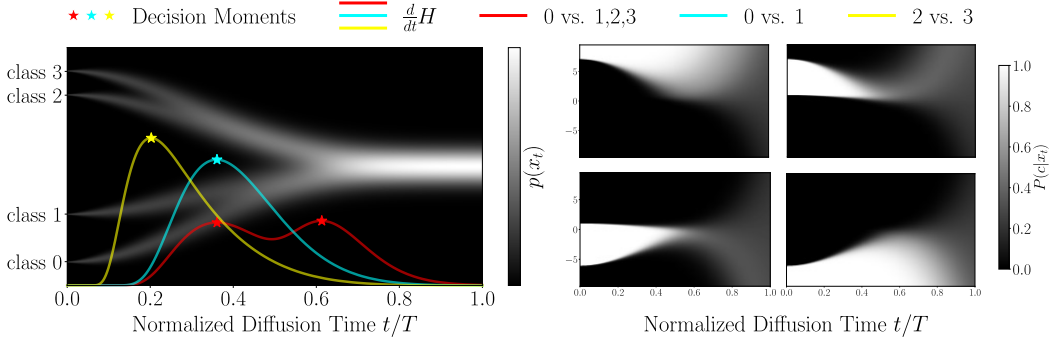
Figure 1: \textbf{Generative information transfer in 1D Gaussian-mixture diffusion}. \label{fig: gmm_diffusion}\\
\textbf{Left}) Diffusion of a one-dimensional equally weighted Gaussian-mixture with four data points (classes) at (-8,-4, 6, 8) along with $\dot{H}(\rz|\rx_t)$ for different decision problems. \textbf{Right}) Posterior evolutions of classes 3 (top left), 2 (top right), 1 (bottom left), 0 (bottom right). The time axes represent the normalized number of noise additions according to a linear noising schedule.  

The main idea is to measure the generative information production relative to a defined decision problem by computing the conditional entropy of the accordingly partitioned classes given the noisy states. The importance of a given time point for the generation of a given class can then be assessed through the corresponding entropy production (i.e., the temporal derivative of the conditional entropy) which quantifies how many bits of information have been transferred by the denoising process relative to the defined partition.

\section{Generative Diffusion from an information theoretical perspective}\label{cond_entr}
The amount of information retained in a data source can be quantified by the entropy and is typically measured in bits \citep{Shannon1948}:
\begin{equation*}
H(\ry) = -\sum_{j=1}^{n} P(y_j)\log_2 P(y_j)  
\end{equation*}
where $y_j$ are the values of a random variable $\ry$ defined on vocabulary $\mathcal{X}$ with $n$ elements and $P(\rvy)$ is its probability mass function. In generative diffusion, the marginal entropy $H(\rx_t)$ can be used for addressing temporal questions, such as the generative information potential of a model at a specific point in time. However, it does not quantify the semantic content of the information that is restored. By shifting the focus to the class label random variable $\rc$, conditional to the state random variable $\rx_t$, we can derive a formulation that quantifies semantic information.\\

% figure 2
\begin{tikzpicture}[node distance=4cm]
    \node (source) [block, minimum width=0.5cm, minimum height=1.5cm, text width=2cm] {Source \\ $p(x_0, c)$};
    \node (transmitter) [block, right of=source, minimum width=0.5cm, minimum height=1.5cm, text width=2.1cm] {Transmitter \\ $p(x_{t}|x_0,c)$};
    \node (receiver) [block, right of=transmitter, minimum width=0.5cm, minimum height=1.5cm, text width=2.1cm] {Receiver \\ $P(c|x_t)$};
    \node (destination) [block, right of=receiver, minimum width=0.5cm, minimum height=1.5cm, text width=2.1cm] {Destination \\ $P(c)$};
    \draw [arrow] (source) -- node[anchor=south] {$x_0, c$} (transmitter);
    \draw [arrow] (transmitter) -- node[anchor=south] {$x_t$} (receiver);
    \draw [arrow] (receiver) -- node[anchor=south] {$c$} (destination);
\end{tikzpicture}

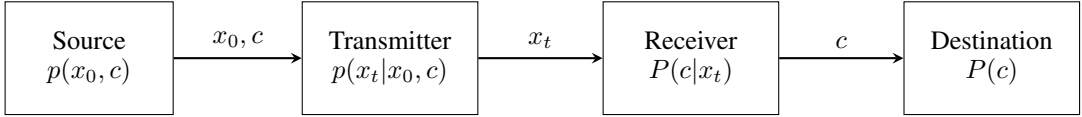
Figure 2: \textbf{Transmitting class-semantic information in generative diffusion}.\label{fig: diffusion_inf_diagram}\\
Information transmission in generative diffusion from the perspective of the class random variable $\rc$. In contrast to the conventional description, the receiver $p(x_0|x_t)$ is exchanged with a classifier $P(c|x_t)$, and the destination $p(x_0)$ is replaced by $P(c)$. 

As illustrated in Figure 2, the class generation process can be conceptualized as a form of information transfer through a noisy channel given by the forward process. An image $x_0$ is sampled from the dataset and then transmitted through a noisy channel $x_0 \rightarrow x_T$ determined by the forward process, where a decoder recovers information concerning its class label. The optimal probabilistic decoder can be obtained using Bayes rule $P(c_k | x_t) = \frac{p(x_t | c_k) P(c_k) }{\sum_{j=1}^n p(x_t | c_j) P(c_j)}$.
The uncertainty in the class label random variable $\rc$ given a noisy state random variable $\rx_t$ may then be quantified through the conditional entropy
\begin{equation}
    H(\rc| \rx_t) = - \int p(x_t) \sum_{k=1}^K P(c_k|x_t) \log_2 P(c_k|x_t) \ dx_t \label{eq: cond_entropy}
\end{equation}
%%
%which quantifies the expected uncertainty left in $\rc$ when knowing $\rx_t$.
We can thus express the total amount of information transmitted up to time $t$ as $T_t = H(\rc) - H(\rc| \rx_t)$.
This \emph{information transfer} captures the amount of information that has been gained by measuring the noisy state up to time $t$. Note that this information is fully contained in the current state $x_t$ since in a Markov forward process we have $P(c | x_{t:T}) = P(c | x_{t})$. The specific contribution of the (infinitesimal) noise added at time $t$ can be quantified by the entropy rate $\dot{H}(\rc | \rx_t)$, which is the temporal derivative of the conditional entropy.

\section{The conditional entropy}
The conditional entropy, as defined in Eq. \ref{eq: cond_entropy}, encompasses the entire set of possible classes. We introduce a binary variable $\rz \in \{z_0, z_1\}$ to partition this set leading to the following definition. 
\begin{equation}
    H(\rz|\rx_t) = - \int p_{\rz}(x_t) \sum_{z} P(z|x_t) \log_2 P(z|x_t)dx_t \label{eq: cond_entropy_bin}
\end{equation}
Here, $p_{\rz}(x_t)$ denotes the mixture distribution under the partition induced by $\rz$. The prior $P(z)$ can be computed from the corresponding class-specific priors. When equal to 0.5, Eq. \ref{eq: cond_entropy_bin} becomes proportional to the negative Jensen-Shannon Divergence between the mixture components $p_{z_0}$ and $p_{z_1}$ (Appendix \ref{appendix: jsd_cond_entr}). Eq. \ref{eq: cond_entropy_bin} measures the uncertainty left in the decision defined by $\rz$ given $\rx_t$. When defining $z_0$ to represent one specific class and $z_1$ to represent everything else, we can measure class-specific generative information transfer. Figure \ref{fig: gmm_diffusion} shows the temporal derivative of Eq. \ref{eq: cond_entropy_bin} for different decision problems. We can see that the maxima of the temporal derivatives align with the bifurcations in the diffusion process of the branches that are involved in the respective decisions. For Gaussian-mixture models, the conditional entropy can be approximated readily as the joint distributions $p(x_t, c)$ are Gaussian, with diffused class-specific mean and variance $\mu_{kt}, \sigma^2_{kt}$, weighted by a class-specific prior $\pi_k$ (Appendix \ref{appendix: gmm_diff}). 

%We can thus analytically evaluate the posteriors $P(z|x_t)$. The expectation with respect to $p_\rz(x_t)$ still needs to be approximated which can be done easily in 1D using the Riemann sum. However, when applying our approach to models trained on complex datasets with unknown initial distributions the posterior is no longer available in closed form. 

\subsection{Estimating the entropy in trained models} \label{post_approx}
To estimate the entropy in trained models, we use the algorithm proposed by \citet{koulischer2025dynamicnegativeguidancediffusion} which estimates the class-specific posterior $P(c|x_t)$ by tracking the Markov chain of the forward process backward in time, i.e., $p(x_{T:t}) = p(x_T)\prod_{\tau=T}^{t+1}p(x_{\tau-1}|x_\tau)$. Using the data processing inequality which directly follows from the Markovian structure of the forward process, they show that the likelihood of a specific class depends only on the least noisy state, i.e. $P(c|x_{T:t}) = P(c|x_{t})$. Thus, the posterior can be estimated iteratively for the current denoising trajectory:   
\begin{equation}
\begin{split}
    \log P(c|x_t) 
    &= 
    \log P(c|x_{t+1}) + \big(\log p(x_t|x_{t+1},c)-\log p(x_t|x_{t+1})\big)\\
    &= 
    \log P(c|x_{t+1}) - \frac{1}{2\sigma_t^2}\big(||x_t-\mu_{\theta}(x_{t+1};c)||^2-||x_t-\mu_\theta(x_{t+1})||^2\big) \label{eq: post_approx}
    \end{split}
\end{equation}
The means, $\mu_{\theta}(x_{t+1};c)$ and $\mu_\theta(x_{t+1})$ can be obtained from a model's conditional and unconditional noise predictions respectively. In its present form, Eq. \ref{eq: post_approx} approximates the posterior ratio between a specific class and "everything" which is encoded by the unconditional model, also referred to as \textit{null}-model $p_\theta(x_t|\emptyset)$ in classifier-free guidance literature \citep{ho2022classifierfreediffusionguidance}. We redefine Eq. \ref{eq: post_approx} in terms of the binary variable $\rz$. When partitioning the set of classes into a specific class versus everything else, Eq. \ref{eq: post_approx} yields both, $P(z_0|x_t)$ and $P(z_1|x_t) = 1 - P(z_0|x_t)$. We sample from the mixture distribution $p_\rz(x_t)$ by splitting the expectation in Eq. \ref{eq: cond_entropy_bin} into its mixture components and then using the learned scores, $\nabla_{x_t}\log p_\theta(x_t|z_0)$ and $\nabla_{x_t} p_\theta(x_t|z_1)\big)$. When comparing one class with its complement, we choose to approximate $\nabla \log p(x_t|\neg c)$ by the \textit{null}-model, i.e., $\nabla_{x_t}\log p_\theta(x_t|\emptyset)$ as it allows us to estimate Eq. \ref{eq: cond_entropy_bin} using two forward passes through a trained model for every sample $x_t$. The larger the set of classes, the better this approximation becomes. By exchanging the \textit{null}-model with a second class-specific model, we can directly compare two classes. A more detailed description can be found in Appendix \ref{appendix: cifar_diff}.

\section{Generative Information Transfer in CIFAR10}
Figure 3 shows the estimated temporal derivatives of the conditional entropy for the decision problems just described for a model trained on CIFAR10 (Appendix \ref{appendix: impl_det}). Here, we use "deer" as the target class ($z_0$) and "airplane", "bird", "cat", "car", and "not deer" as the counterparts ($z_1$) respectively. 

% figure 3
\begin{center}
    \includesvg[width=0.63\linewidth]{figures/deer.svg}
\end{center}
Figure 3: \textbf{Generative information transfer in CIFAR10}.\label{fig: gen_inf_transfer_cifar} \\
\textbf{Left}) Expected noiseless images throughout the generative process for the five example classes "deer", "airplane", "bird", "cat", "car", and the "null" class which we use to approximate "not deer". \textbf{Right}) Approximated temporal derivatives for the binary decisions between "deer" and "airplane", "bird", "cat", "car", and "not deer", respectively.

The temporal derivatives of the conditional entropy of decision problems involving contrasting classes, e.g., "deer" versus "car," which supposedly share minimal structural information, peak at an earlier stage in the generative process, suggesting a preliminary divergence of the respective diffusion branches, similar to the bifurcations illustrated in Figure 1. However, we also observe that across experiments, including those in Appendix \ref{appendix: cifar_diff}, most of the information transfer occurs within a central interval that is preceded by a data-mean convergence, emphasizing the existence of different diffusion regimes and a \textit{critical window} \citep{raya2023spontaneoussymmetrybreakinggenerative, ambrogioni2024statisticalthermodynamicsgenerativediffusion, li2024criticalwindowsnonasymptotictheory, Biroli_2024}. 

\section{Conclusion \& Broader Impact}
Although we have outlined a comprehensive method for measuring class-semantic information transfer in generative diffusion models, it still remains to be extensively validated on models trained on more complex class-conditional datasets, such as ImageNet, and those trained on datasets with compositional class hierarchies, such as LAION. In addition, we will attempt to derive an analytical connection between the conditional entropy and the critical decision points corresponding to the first-order phase transitions described by \citet{raya2023spontaneoussymmetrybreakinggenerative}. Nevertheless, even in its current form, our approach provides an alternative view of generative diffusion that facilitates our theoretical understanding of the topic. Additionally, it offers a theoretically sound method for performing guided conditional generation in a semi-dynamic fashion that could be the starting point to extend the ideas from \citet{kynkäänniemi2024applyingguidancelimitedinterval} and improve their practical applicability.

\newpage
\bibliography{main}
\bibliographystyle{iclr2025_delta}
\newpage
\appendix

% Appendix A
\section{Jensen-Shannon Divergence and Conditional Entropy} \label{appendix: jsd_cond_entr}
The Jensen-Shannon Divergence ($JSD$) can be defined using Kullback-Leibler Divergences between two distributions $p$, $q$ and their mixture $m=\frac{p+q}{2}$. With $p=p(x_t|z_0)$ and $q=p(x_t|z_1)$ $JSD$ can accordingly be written as:
\begin{align*}
    \begin{split}
        JSD(p(x_t|z_0)||p(x_t|z_1)) 
        &= 
        \frac{1}{2} \big(D_{KL}\big(p(x_t|z_0)||m\big)+D_{KL}\big(p(x_t|z_1)||m\big)\big)\\
        &= 
        \frac{1}{2} \int p(x_t|z_0)\log\bigg(\frac{2\cdot p(x_t|z_0)}{p(x_t|z_0)+p(x_t|z_1)}\bigg) \\
        &\qquad + p(x_t|z_1)\log\bigg(\frac{2\cdot p(x_t|z_1)}{p(x_t|z_0)+p(x_t|z_1)}\bigg) \ dx_t\\
        &= 
        \frac{1}{2} \int p(x_t|z_0)\log\bigg(\frac{p(x_t|z_0)}{p(x_t|z_0)+p(x_t|z_1)}\bigg)\\
        &\qquad + p(x_t|z_1)\log\bigg(\frac{p(x_t|z_1)}{p(x_t|z_0)+p(x_t|z_1)}\bigg) \ dx_t + \log 2
    \end{split}    
\end{align*}
Now, $\log (\frac{2p}{m})$ can be written in terms of the posterior and prior: 
\begin{equation*}
    \log\bigg(\frac{p(x_t|z_0)}{p(x_t|z_0)+p(x_t|z_1)}\bigg) =  \log\bigg(\frac{P(z_0|x_t)/P(z_0)}{P(z_0|x_t)/P(z_0)+P(z_1|x_t)/P(z_1)}\bigg)\\
\end{equation*}
which we will denote as $\log p$ for convenience from now on as well as $\log 2 := c$: 
\begin{align*}
    \begin{split}
        JSD(p(x_t|z_0)||p(x_t|z_1)) 
        &= 
        \frac{1}{2} \int p(x_t|z_0)\cdot\log p + p(x_t|z_1)\cdot\log (1-p) \ dx_t + c\\
        &= 
        \frac{1}{2} \int \frac{P(z_0|x_t)p(x_t)}{P(z_0)}\cdot\log p + p(x_t|z_1)\cdot\log (1-p) \ dx_t + c\\
        &= 
        \frac{1}{2} \int \frac{P(z_0|x_t)/P(z_0)+P(z_1|x_t)/P(z_1)}{P(z_0|x_t)/P(z_0)}\\
        &\qquad +
        P(z_1|x_t)/P(z_1)\cdot \frac{P(z_0|x_t)}{P(z_0)}p(x_t)\cdot \log p \\
        &\qquad + 
        p(x_t|z_1)\cdot\log (1-p) \ dx_t + c\\
        &= 
        \frac{1}{2} \int \bigg[\frac{P(z_0|x_t)p(x_t)}{P(z_0)}+\frac{P(z_1|x_t)p(x_t)}{P(z_1)}\bigg]\cdot p\log p \\
        &\qquad + 
        p(x_t|z_1)\cdot\log (1-p) \ dx_t + c\\
        &= 
        \frac{1}{2} \int \bigg[p(x_t|z_0)+p(x_t|z_1)\bigg]\cdot p\log p + p(x_t|z_1)\cdot\log (1-p) \ dx_t + c
    \end{split}
\end{align*}
Likewise, one can transform the right summand: 
\begin{equation*}
    p(x_t|z_1)\cdot\log (1-p) = \bigg[p(x_t|z_0)+p(x_t|z_1)\bigg] (1-p) \log(1-p)
\end{equation*}
The final result then yields: 
\begin{align*}
\begin{split}
JSD(p(x_t|z_0)||p(x_t|z_1)) 
&= \frac{1}{2} \int \bigg[p(x_t|z_0)+p(x_t|z_1)\bigg]\cdot p\log p \\ 
&\qquad + \bigg[p(x_t|z_0)+p(x_t|z_1)\bigg] \cdot (1-p) \log(1-p) \ dx_t+ c\\
&= \frac{1}{2} \int p(x_t|z_0) \cdot \bigg[p\log p + (1-p)\log(1-p)\bigg] \ dx_t  \\ 
&\qquad + \frac{1}{2} \int p(x_t|z_1) \cdot \bigg[p\log p  + (1-p)\log(1-p)\bigg] \ dx_t+ c\\
&= \int \frac{p(x_t|z_0)+p(x_t|z_1)}{2} \cdot \bigg[p \log p + (1-p)\log(1-p)\bigg] dx_t + c
\end{split}
\end{align*}

% Appendix B
\section{Gaussian Mixture Diffusion} \label{appendix: gmm_diff}
\subsection{Deriving the marginals, likelihoods, and posteriors}
When diffusing a Gaussian-mixture model (GMM) the intermediate distributions are available in closed-form. We will derive them in the following. Starting from the definition of a GMM
\begin{equation*}
    p(x_0) = \sum_k^K \pi_k\mathcal{N}(x_0|\mu_k, \sigma^2_k)
\end{equation*}
and applying a variance-preserving noise addition with known $\beta_t$, the diffusion kernel is defined as: 
\begin{equation*}
    p(x_t|x_0) = \mathcal{N}(x_t|\sqrt{\alpha_t}x_0, 1-\alpha_t) \text{ with } \alpha_t = \prod_{\tau=1}^t(1-\beta_t)
\end{equation*}
Accordingly, the marginals are given by: 
\begin{equation*}
        p(x_t) = \sum_k^K \pi_k \int \mathcal{N}(x_t|\sqrt{\alpha_t}x_0, 1-\alpha_t)\cdot \mathcal{N}(x_0|\mu_k, \sigma^2_k) \ dx_0
\end{equation*}
The integral resolves to another Gaussian:
\begin{equation*}
    I= \int \mathcal{N}(x_t|\sqrt{\alpha_t}x_0, 1-\alpha_t)\cdot \mathcal{N}(x_0|\mu_k, \sigma^2_k) \ dx_0 = \mathcal{N}(x_t|\mu_{kt}, \sigma^2_{kt})
\end{equation*}
With $\mu_{kt}$ and $\sigma^2_{kt}$ derived in the following: 
\begin{equation*}
\begin{split}
    I 
    &= 
    \frac{1}{\sqrt{2\pi\sigma_k^2}}\frac{1}{\sqrt{2\pi(1-\alpha_t)}}\int \exp\bigg(E\bigg)dx_0\\
    E 
    &= 
    -\frac{1}{2\sigma_k^2}(x_0-\mu_k)^2-\frac{1}{2(1-\alpha_t)}(x_t-\sqrt{\alpha_t}x_0)^2\\
    &=
    -\frac{x_0^2}{2\sigma_k^2}+\frac{x_0\mu_k}{\sigma_k^2}-\frac{\mu_k^2}{2\sigma_k^2}-\frac{x_t^2}{2(1-\alpha_t)}+\frac{\sqrt{\alpha_t}x_0}{1-\alpha_t}-\frac{\alpha_tx_0^2}{2(1-\alpha_t)}\\
    &= 
    -\frac{1}{2}\bigg(\frac{1}{\sigma_k^2}+\frac{\alpha_t}{1-\alpha_t}\bigg)x_0^2+\bigg(\frac{\mu_k}{\sigma_k^2}+\frac{x_t\sqrt{\alpha_t}}{1-\alpha_t}\bigg)x_0-\frac{\mu_k^2}{2\sigma_k^2}-\frac{x_t^2}{2(1-\alpha_t)}\\
    &= 
    -\frac{A}{2}x_0^2+Bx_0+C = -\frac{A}{2}(x_0-\frac{B}{A})+\frac{B^2}{2A}+C\\
    &\Rightarrow\\
    I 
    &= \frac{1}{\sqrt{2\pi\sigma_k^2}}\frac{1}{\sqrt{2\pi(1-\alpha_t)}}\exp\bigg(\frac{B^2}{2A}+C\bigg) \int \exp\bigg(-\frac{A}{2}(x_0-\frac{B}{A})^2\bigg)dx_0\\
    &= 
    \frac{1}{\sqrt{2\pi\sigma_k^2}}\frac{1}{\sqrt{2\pi(1-\alpha_t)}}\exp\bigg(\frac{B^2}{2A}+C\bigg) \frac{\sqrt{2\pi}}{\sqrt{A}}\\
    &= 
    \frac{1}{\sqrt{2\pi\sigma^2_k(1-\alpha_t)A}}\exp\bigg(\frac{B^2}{2A}+C\bigg)\\
\end{split}
\end{equation*}
We can directly read off $\sigma^2_{kt}$:
\begin{equation*}
\begin{split}
    \sigma^2_{kt} 
    &= 
    \sigma_k^2(1-\alpha_t)A = \alpha_t\sigma^2_k+(1-\alpha_t)\\
\end{split}
\end{equation*}
Likewise, we recognize that
\begin{equation*}
    \begin{split}
    \frac{B^2}{A} + C 
    &= -\frac{1}{2\sigma_{kt}^2}x_t^2-\frac{\mu_{kt}}{\sigma_{kt}^2}x_t + \frac{1}{2\sigma_{kt}^2}c\\
    \frac{\mu_{kt}}{\sigma_{kt}^2}x_t 
    &= 
    \frac{2(\mu_k\sqrt{\alpha_t}) / (\sigma_k^2(1-\alpha_t))}{2((1-\alpha_t)+\alpha_t\sigma_k^2) / (\sigma_k^2(1-\alpha_t))}x_t = \frac{\mu_k\sqrt{\alpha_t}}{\sigma^2_{kt}}x_t\\
\end{split}
\end{equation*}
Again, we can read off $\mu_{kt}$: 
\begin{equation*}
\begin{split}
    \mu_{kt} 
    &= 
    \sqrt{\alpha_t}\mu_k
    \end{split}
\end{equation*}
The likelihood, $p(x_t|c)$ is then simply given by $\mathcal{N}(x_t|\mu_{kt}, \sigma^2_{kt})$. The only term left is the posterior $P(c|x_t)$ that can be derived from the previous definitions and Bayes' formula: 
\begin{equation*}
    P(c_k|x_t) = \frac{p(x_t|c_k)P(c_k)}{p(x_t)} = \frac{\pi_k\mathcal{N}(x_t|\mu_{kt}, \sigma_{kt}^2)}{\sum_j^K \pi_j\mathcal{N}(x_t|\mu_{jt}, \sigma_{jt}^2)}
\end{equation*}

\subsection{Estimating the Conditional Entropy}
Recall the definition of the conditional entropy defined under the binary decision $\rz$: 
\begin{equation*}
    H(\rz|\rx_t) = -\int p_\rz(x_t) \sum_z P(z|x_t) \log P(z|x_t) dx_t
\end{equation*}
Consider the decision between two specific classes, $z_0 \rightarrow c_1$, and $z_1 \rightarrow c_2$ that are part of a larger set $\rc = \{c_1, c_2, c_3, c_4\}$. We can then write $p_\rz(x_t)$  
\begin{equation*}
    p_\rz(x_t) = p(x_t|z_0)P(z_0)+p(x_t|z_1)P(z_1)
\end{equation*}
where the priors are defined relative to the decision problem, i.e., $P(z_0) = P(c_1)/(P(c_1)+P(c_2))$ and $P(z_1)=P(c_2)/(P(c_1)+P(c_2))$. Similarly, the posteriors are defined as: 
\begin{equation*}
\begin{split}
    P(z_0|x_t) &= \frac{P(c_1|x_t)}{P(c_1|x_t)+P(c_2|x_t)}\\
    P(z_1|x_t) &= \frac{P(c_2|x_t)}{P(c_1|x_t)+P(c_2|x_t)}
\end{split}
\end{equation*}
The likelihoods are given by $p(x_t|z_0) = p(x_t|c_1)$ and $p(x_t|z_1)=p(x_t|c_2)$. Thus the conditional entropy for the given decision problem becomes:
\begin{align*}
\begin{split}
    H(\rz|\rx_t) 
    &= 
    -\int \bigg[p(x_t|c_1)\frac{P(c_1)}{P(c_1)+P(c_2)} + p(x_t|c_2)\frac{P(c_2)} {P(c_1)+P(c_2)}\bigg] \cdot\\
    &\qquad \cdot 
    \sum_j^2\frac{P(c_j|x_t)}{P(c_1|x_t)+P(c_2|x_t)} \log \bigg(\frac{P(c_j|x_t)}{P(c_1|x_t)+P(c_2|x_t)}\bigg) dx_t\\
    &= -\sum_i^2 \frac{P(c_i)}{P(c_1)+P(c_2)}\int p(x_t|c_i)\sum_j^2\frac{P(c_j|x_t)}{P(c_1|x_t)+P(c_2|x_t)} \log \bigg(\frac{P(c_j|x_t)}{P(c_1|x_t)+P(c_2|x_t)}\bigg) dx_t\\
\end{split}
\end{align*}
The likelihoods $p(x_t|c_i)$ and the posteriors $P(c_j|x_t)$ were derived in the previous section and are given by: 
\begin{equation*}
    \begin{split}
        p(x_t|c_i) &= \mathcal{N}(x_t|\mu_{it}, \sigma_{it}^2)\\
        P(c_j|x_t) &= \frac{\pi_j\mathcal{N}(x_t|\mu_{jt}, \sigma_{jt}^2)}{\pi_1\mathcal{N}(x_t|\mu_{1t}, \sigma_{1t}^2)+\pi_2\mathcal{N}(x_t|\mu_{2t}, \sigma_{2t}^2)}
    \end{split}
\end{equation*}
The priors are simply $P(c_i)=\frac{\pi_i}{\pi_1+\pi_2}$. We estimate the integral using a discretization for $x_t$ and then taking the Riemann sum. 

\subsection{Additional Examples: Diffusion and Fixed-point Dynamics}
Additional setups for symmetric and asymmetric GMMs, equally weighted as well as differently weighted. Additionally, we show the corresponding fixed point dynamics that were derived by optimizing $-\nabla_{x_t}\log p(x_t)+0.5x_t$ using Powell's hybrid method (Newton-Raphson + steepest descent).

\includesvg[width=1\linewidth]{appendix/2deltas_1.svg}

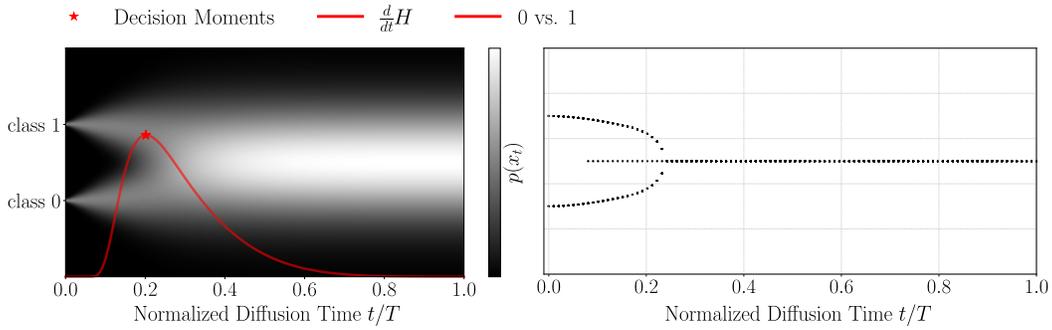
Figure 4: GMM with mixture weights $(\pi_0, \pi_1) = (0.5,0.5)$ and deltas at $(\mu_0, \mu_1) = (-1,1).$ 

\includesvg[width=1\linewidth]{appendix/2deltas_2.svg}

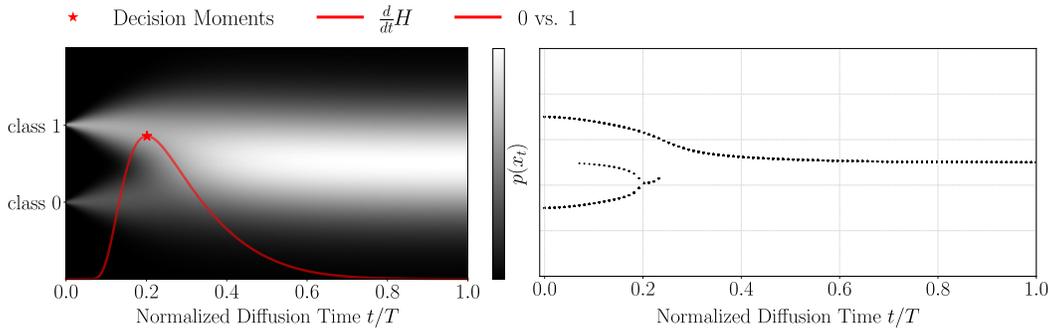
Figure 5: GMM with mixture weights $(\pi_0, \pi_1) = (1/3,2/3)$ and deltas at $(\mu_0, \mu_1) = (-1,1).$ 

\includesvg[width=1\linewidth]{appendix/3deltas_1.svg}

Figure 6: GMM with mixture weights $(\pi_0, \pi_1, \pi_2) = (1/3,1/3,1/3)$ and deltas at $(\mu_0, \mu_1, \mu_2) = (-2,0,2).$ 

\includesvg[width=1\linewidth]{appendix/3deltas_2.svg}

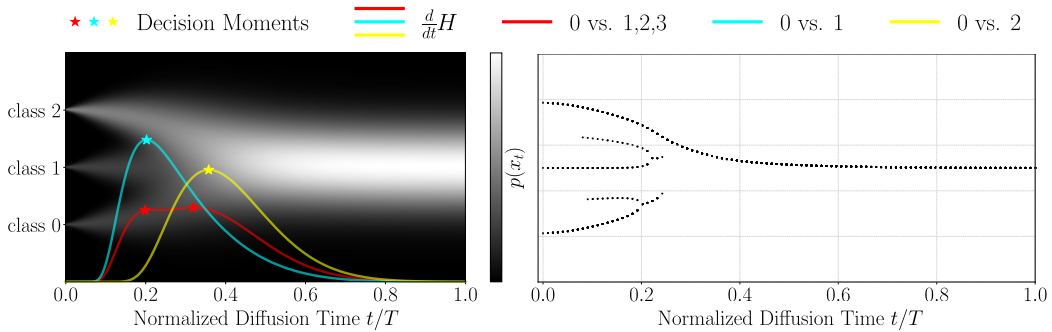
Figure 7: GMM with mixture weights $(\pi_0, \pi_1, \pi_2) = (0.25,0.25,0.5)$ and deltas at $(\mu_0, \mu_1, \mu_2) = (-2,0,2).$ 

\includesvg[width=1\linewidth]{appendix/3deltas_3.svg}

Figure 8: GMM with mixture weights $(\pi_0, \pi_1, \pi_2) = (1/3,1/3,1/3)$ and deltas at $(\mu_0, \mu_1, \mu_2) = (-2,1,2).$ 

\includesvg[width=1\linewidth]{appendix/4deltas_1.svg}

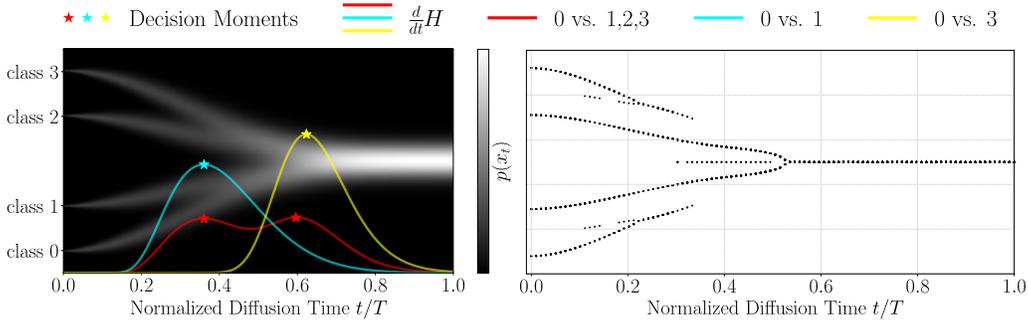
Figure 9: GMM with mixture weights $(\pi_0, \pi_1, \pi_2, \pi_3) = (0.25,0.25, 0.25, 0.25)$ and deltas at $(\mu_0, \mu_1, \mu_2, \mu_3) = (-8,-4,4, 8).$ 

% % Appendix D
% \section{Posterior Approximation For Arbitrary Decision Problems}
% When defining $z$ s.t. it does not* 

% Appendix D
\section{Diffusion on CIFAR10} \label{appendix: cifar_diff}
\subsection{Estimating the Conditional Entropy}
To estimate Eq. \ref{eq: cond_entropy_bin} we need to approximate the sampler, $p_{z}(x_t)$ and the posterior, $P(z_0|x_t)$. We split the integral in Eq. \ref{eq: cond_entropy_bin} into two parts: 
\begin{equation*}
- \sum_z P(z)\int p(x_t|z) \bigg[P(z_0|x_t) \log P(z_0|x_t)+P(z_1|x_t) \log P(z_1|x_t) \bigg] dx_t
\end{equation*}
We obtain $N_{z_0}=N_{z_1}=1000$ samples for the individual expectations using ancestral sampling (T=1000 inference steps)
\begin{equation*}
   x_{t-1}^z = \frac{1}{\sqrt{1-\beta_t}}(x_t^z-\frac{\beta_t}{\sqrt{1-\alpha_t}} \epsilon_\theta(x_{t};z)\big) + \sqrt{\beta_t}\epsilon\\
\end{equation*}
where $\epsilon_\theta(x_t;z) = - \sqrt{1-\alpha_t}\nabla_{x_t}\log p(x_t|z)$. For $c$ vs.$ \neg c$, we use $\nabla_{x_t}\log p(x_t|\neg c) \approx \nabla_{x_t}\log p(x_t|\emptyset)$. We estimate the posteriors $P(z_0|x_t) = 1 - P(z_1|x_t)$ using the iterative procedure from \citet{koulischer2025dynamicnegativeguidancediffusion}: 
\begin{equation*}
    P(z_0|x_{t-1}) = P(z_0|x_{t})\exp\big(\frac{1}{1-\beta_t}\big(||x_{t-1} - \mu_\theta(x_{t};z_0)||^2 - ||x_{t-1}-\mu_\theta(x_{t};z_1)||^2\big)\big)
\end{equation*}
The individual expectations are weighed by their respective priors, $P(z)$: $(0.5, 0.5)$ for $c_1$ vs. $c_2$ and $(0.1, 0.9)$ for $c$ vs. $\neg c$. Here is the procedure outlined in more detail:  
\begin{algorithm}
\caption{Estimate Conditional Entropy}
\begin{algorithmic}[1] 
\State \textbf{Input:} $N_{z_0}, N_{z_1}, T, P(z_0)$
\State $x_T^{z_0}$, $x_T^{z_1} \sim \mathcal{N}(0,I)$
\State$P(z_0|x_T) =  P(z_0)$
\State $H_T= - \big(P(z_0)\log P(z_0) + (1-P(z_0))\log (1-P(z_0))\big)$
\For{$t$ in $T:1$}
    \For{$z \in \{z_0, z_1\}$ } 
    \State $x_t = x_t^z$
    \State $\mu_\theta(x_t; z_0) = \frac{1}{\sqrt{1-\beta_t}}\big(x_t-\frac{\beta_t}{\sqrt{1-\alpha_t}}\epsilon_\theta(x_t,z_0)\big)$ 
    \State $\mu_\theta(x_t;z_1) = \frac{1}{\sqrt{1-\beta_t}}\big(x_t-\frac{\beta_t}{\sqrt{1-\alpha_t}}\epsilon_\theta(x_t,z_1)\big)$
    \If {$t>1$}
        \State $\epsilon \sim \mathcal{N}(0,I) \cdot \sqrt{\beta_t}$
    \Else 
        \State $\epsilon = 0$
    \EndIf
    \State $x_{t-1} = \mu_\theta(x_t;z_0) + \epsilon$
    \State $P(z_0|x_{t-1}) = P(z_0|x_{t})\exp\big(-\frac{1}{1-\beta_t}\big(||x_{t-1}-\mu_\theta(x_t;z_0)||^2-||x_{t-1}-\mu_\theta(x_t;z_1)||^2\big)\big)$
    \State $P(z_1|x_{t-1}) = 1 - P(z_0|x_{t-1})$
    \State $x_{t-1}^z = x_{t-1}$
    \State $H_{t-1}^z = \frac{1}{N_z}\sum_{x_{t-1}} P(z_0|x_{t-1})\log P(z_0|x_{t-1}) + P(z_1|x_{t-1})\log P(z_1|x_{t-1}) $ 
    \EndFor
\State $H_{t-1} = - \big(P(z_0)H_t^{z_0}+(1-P(z_0))H_t^{z_1}\big)$ 
\EndFor
\State \textbf{output:} $H_{T:0}$

\end{algorithmic}
\end{algorithm}

\subsection{Additional Examples}
\begin{center}
    \includesvg[width=0.73\linewidth]{appendix/cat.svg}

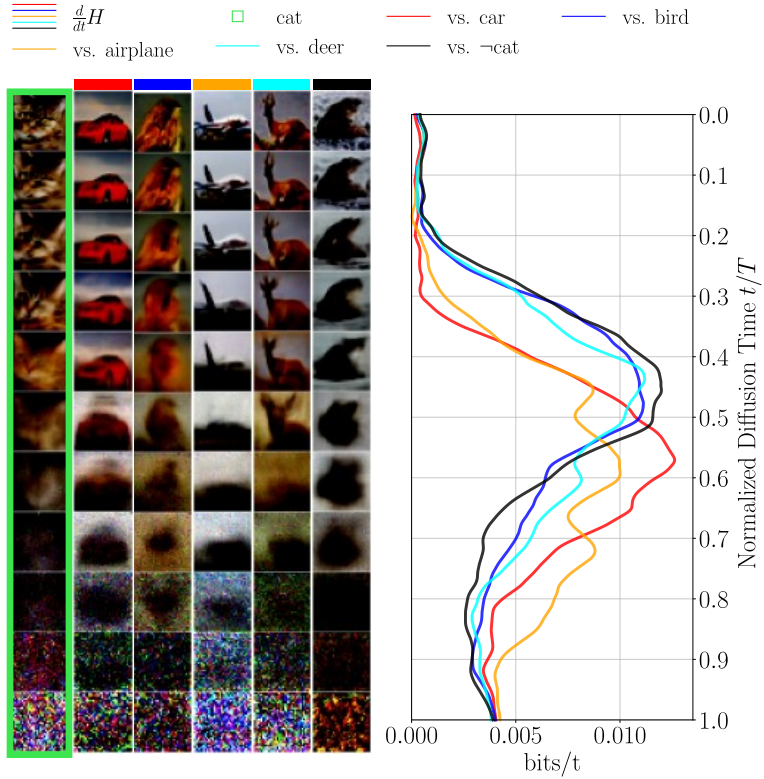
    Figure 10: Experiments using the target class "cat".
\end{center}

\begin{center}
    \includesvg[width=0.73\linewidth]{appendix/airplane.svg}

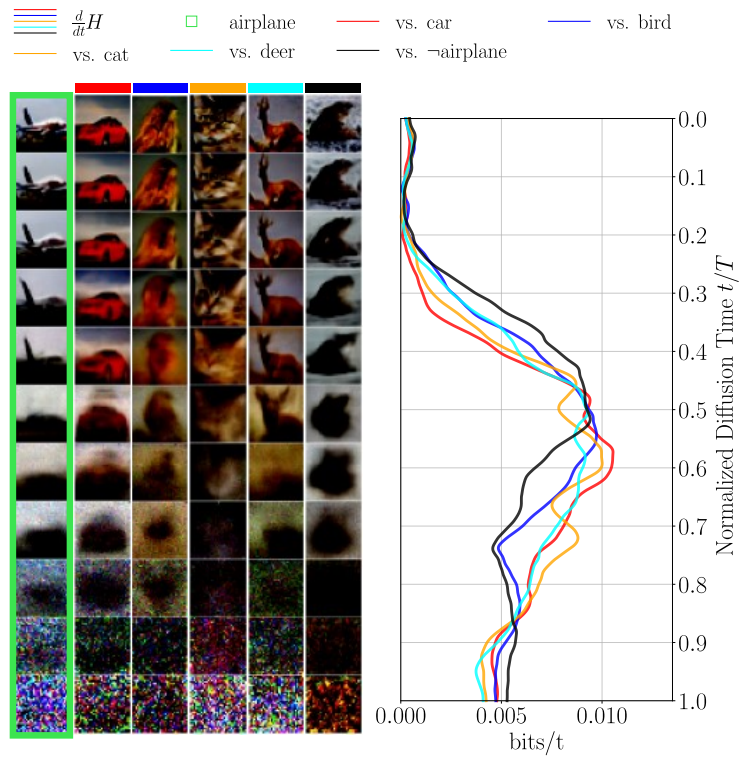
    Figure 11: Experiments using the target class "airplane".
\end{center}

\begin{center}
    \includesvg[width=0.73\linewidth]{appendix/car.svg}

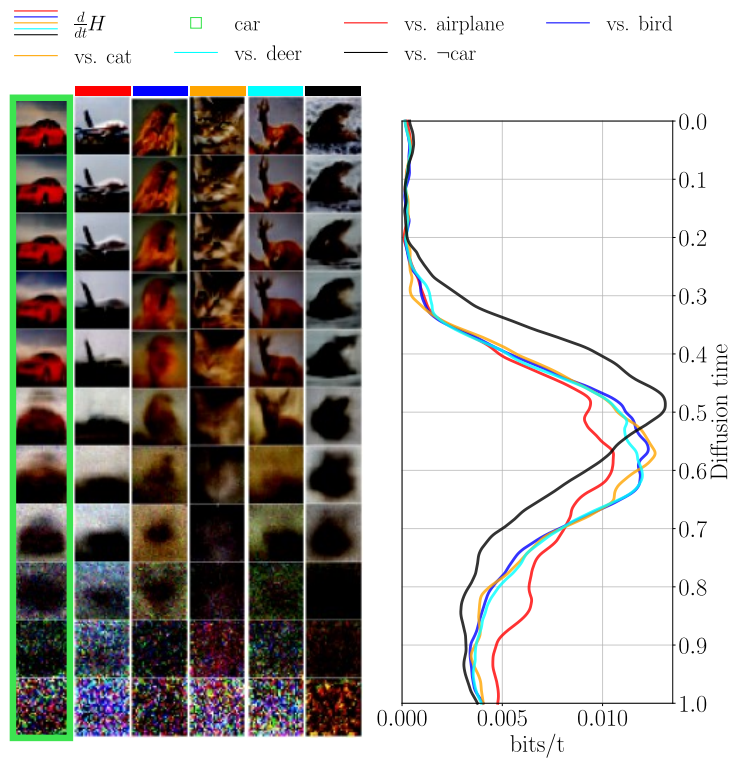
    Figure 12: Experiments using the target class "car".
\end{center}

\begin{center}
    \includesvg[width=0.73\linewidth]{appendix/bird.svg}

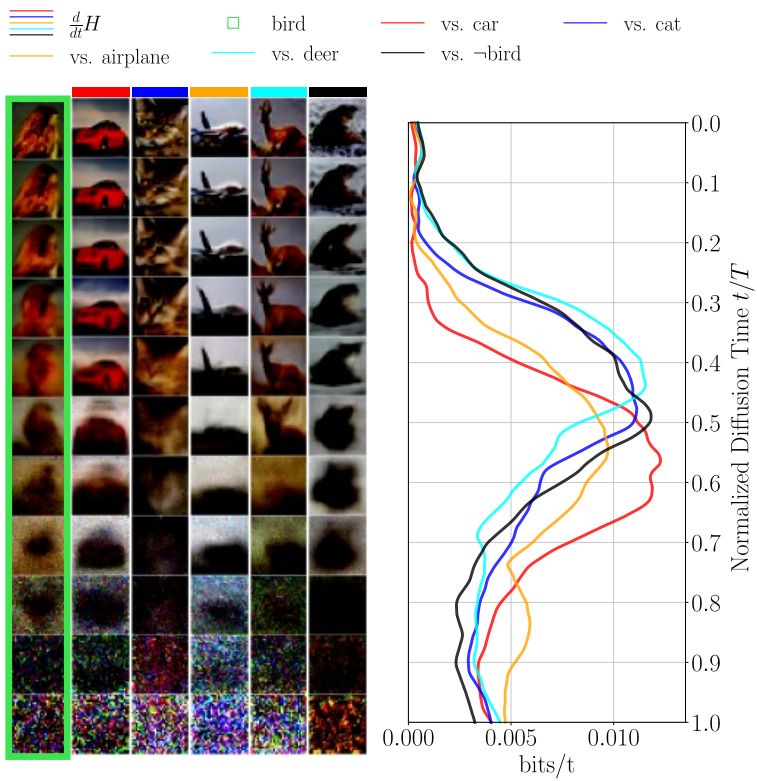
    Figure 13: Experiments using the target class "bird".
\end{center}

\subsection{Implementation Details}\label{appendix: impl_det}

\textbf{Datasets}: All experiments were conducted using the CIFAR-10 \cite{krizhevsky2009learning} dataset, which consists of 50,000 training and 10,000 test images, each image is a $32\times32$ RGB image.  The dataset contains 10 classes: \textit{airplane, car, bird, cat, deer, dog, frog, horse, ship} and \textit{truck}, with 6,000 images per class. We trained two types of models: (1) an unconditional model utilizing the full dataset, and (2) class-specific models for the classes  \textit{airplane, car, bird, cat,} and \textit{deer}, where each model used 6,000 training images for training. 

\textbf{Network architecture}: All models were trained using the U-Net architecture from Denoising Diffusion Probabilistic Models (DDPM) \cite{ho2020denoisingdiffusionprobabilisticmodels}, where the noise prediction model is a modified PixelCNN++ based Unet \cite{van2016conditional, salimans2017pixelcnn++}.

\textbf{Training}: Training was performed using the Adam optimizer with a batch size of 128, distributed across three Tesla V100 GPUs on an NVIDIA DGX-1 machine. The models were trained for 800 epochs using PyTorch 1.10.2+cu102, CUDA 10.2, and CuDNN 7605, with a learning rate of 0.0002, employing random horizontal flips as data augmentation.
\end{document}